\documentclass[sigconf]{acmart}
\usepackage{graphicx}
\usepackage{multirow}
\usepackage{subcaption}
\usepackage{cuted}
\usepackage{caption}
\usepackage{pgfplots}
\pgfplotsset{compat=1.18}
\usepackage{xcolor}
\definecolor{MutedGreen}{RGB}{120,170,130}
\definecolor{MutedRed}{RGB}{190,120,120}
\definecolor{MutedGray}{RGB}{140,140,140}
\usepackage{booktabs}
\usepackage{enumitem}
\usepackage{tikz}
\usepackage{listings}
\usetikzlibrary{arrows.meta, positioning, shapes.geometric, backgrounds, fit, calc}

\definecolor{perf0}{HTML}{C44E52}    
\definecolor{perf1}{HTML}{D4756A}    
\definecolor{perf2}{HTML}{E8A87C}    
\definecolor{perf3}{HTML}{C5B47F}    
\definecolor{perf4}{HTML}{8CB369}    
\definecolor{perf5}{HTML}{5A9367}    
\definecolor{perf6}{HTML}{3D7A5F}    

\title{Building Customer Support AI Agents at 100M-User Scale: An Evaluation-Driven Framework}

\author{Aman Gupta}
\affiliation{%
  \institution{Nubank}
  \city{Palo Alto}
  \country{USA}}
\email{aman.gupta@nubank.com.br}

\author{Kevin Rossell}
\affiliation{%
  \institution{Nubank}
  \city{Mexico City}
  \country{Mexico}}
\email{kevin.rossell@nubank.com.br}

\author{Edesio Alcobaça}
\affiliation{%
  \institution{Nubank}
  \city{São Paulo}
  \country{Brazil}}
\email{edesio.alcobaca@nubank.com.br}

\author{Jose Chrystian Lima Pacheco}
\affiliation{%
  \institution{Nubank}
  \city{São Paulo}
  \country{Brazil}}
\email{chrystian.lima@nubank.com.br}

\author{Carolina Baptista de Lima}
\authornote{Work done while at Nubank.}
\affiliation{%
  \institution{Nubank}
  \city{São Paulo}
  \country{Brazil}}
\email{carolina.bap.lima@gmail.com}

\author{Shao Tang}
\affiliation{%
  \institution{Nubank}
  \city{Palo Alto}
  \country{USA}}
\email{tang.shao@nubank.com.br}

\author{Luiz Paulo Rabachini}
\affiliation{%
  \institution{Nubank}
  \city{São Paulo}
  \country{Brazil}}
\email{luiz.rabachini@nubank.com.br}

\author{Luis Moneda}
\affiliation{%
  \institution{Nubank}
  \city{São Paulo}
  \country{Brazil}}
\email{luis.moneda@nubank.com.br}

\author{Herbert Fei}
\affiliation{%
  \institution{Nubank}
  \city{Palo Alto}
  \country{USA}}
\email{herbert.fei@nubank.com.br}

\author{Daniel Silva}
\affiliation{%
  \institution{Nubank}
  \city{Palo Alto}
  \country{USA}}
\email{daniel.ribeiro@nubank.com.br}

\author{Rohan Ramanath}
\affiliation{%
  \institution{Nubank}
  \city{Palo Alto}
  \country{USA}}
\email{rohan.ramanath@nubank.com.br}

\copyrightyear{2026}
\acmYear{2026}
\setcopyright{cc}
\setcctype{by}
\acmConference[KDD '26]{Proceedings of the 32nd ACM SIGKDD Conference on Knowledge Discovery and Data Mining V.2}{August 09--13, 2026}{Jeju Island, Republic of Korea}
\acmBooktitle{Proceedings of the 32nd ACM SIGKDD Conference on Knowledge Discovery and Data Mining V.2 (KDD '26), August 09--13, 2026, Jeju Island, Republic of Korea}
\acmDOI{10.1145/3770855.3818332}
\acmISBN{979-8-4007-2259-2/2026/08}

\begin{document}

\begin{abstract}
The rapid rise in LLM capabilities has made AI agents increasingly viable across a broad range of tasks. Among the most promising applications is building production-ready customer-facing agents—a challenge that demands coordinated excellence in evaluation methodology, context engineering, training, and online measurement. Yet these critical pillars are typically developed in isolation, creating blind spots that only surface after deployment.

In this paper, we present a unified framework that bridges offline development with online impact for customer support AI agents at Nubank - a company with 100m+ users. Our approach integrates several key components: (1) structured context engineering tailored to customer support agents (2) systematic human-in-the-loop prompt iteration, (3) rigorous LLM judge evaluation with measured inter-rater agreement and GEPA-optimization for consistency and (4) ideation-to-production validation.

A central insight is that evaluation-pipeline quality directly determines iteration velocity. We present results from five production deployments spanning distinct domains—card delivery, debt management, credit-limit support, card management, and product explanation — that deliver consistent customer-satisfaction gains while substantially accelerating iteration. In our card-delivery deployment, large-scale A/B testing yields a 37 percentage-point improvement in AI transactional Net Promoter Score and a 29 percentage-point gain in self-service rate over prior agent variants, alongside a strong correlation between offline simulation metrics and online outcomes—demonstrating that eval-driven development reliably predicts production impact. On most use cases, AI satisfaction reaches within a few percentage points of expert human agents.
\end{abstract}

\renewcommand{\shortauthors}{Aman Gupta et al.}



\maketitle

\section{Introduction}

 The rise of large language models (LLMs)~\cite{brown2020language,dubey2024llama,comanici2025gemini,guo2025deepseek,jiang2023mistral7b} has catalyzed a new paradigm: AI agents capable of dynamically directing their own processes, leveraging tools, and maintaining control over task execution~\cite{anthropic2024agents,yao2023react,jin2025search,feng2025retool}. When equipped with the \textit{right context and tools}, AI Agents have demonstrated impressive capabilities in solving difficult and long running tasks~\cite{kwa2025measuring,erdogan2025planact,chen2025loop}.

 \begin{figure}[t]
    \centering
    \includegraphics[width=0.7\linewidth]{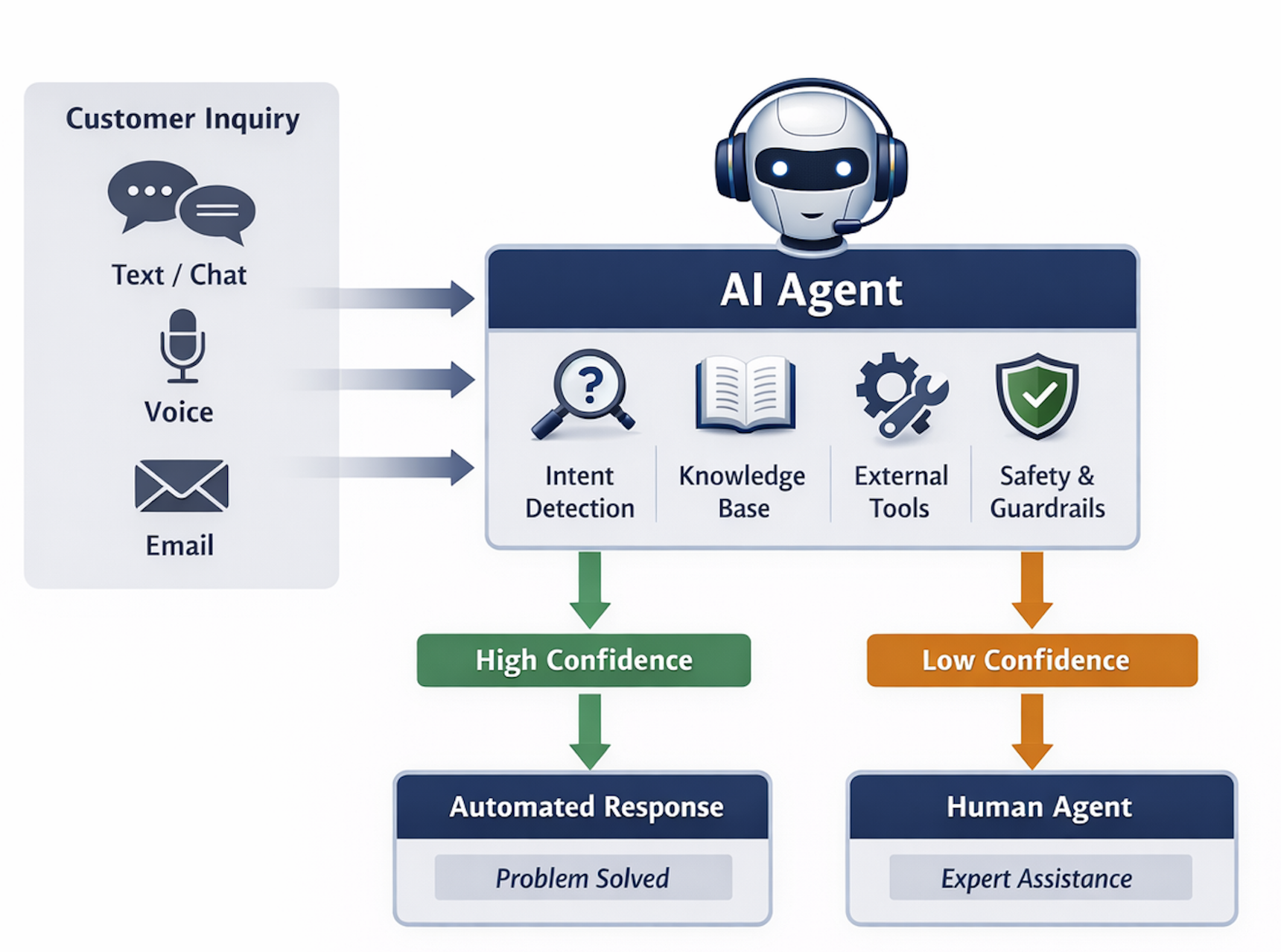}
    \caption{Architecture of a Customer Support AI agent system. User queries arrive via multiple channels and are handled autonomously, with fallback to human agents under low confidence.}
    \label{fig:ai-agent}
\end{figure}

\begin{figure*}[t]
  \centering
  \includegraphics[width=0.9\textwidth]{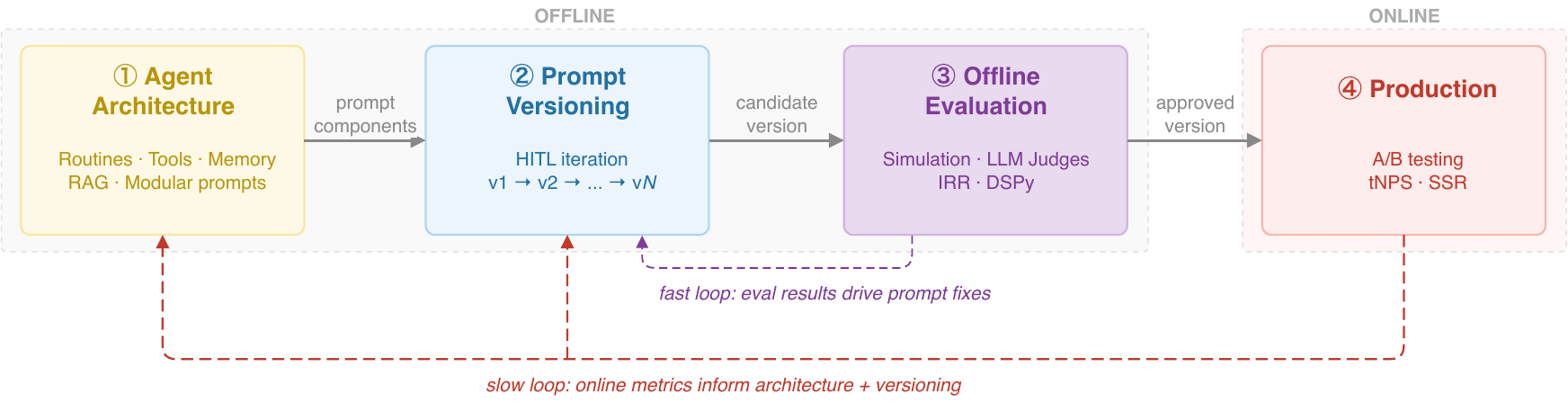}
  \caption{Overview of our evaluation-driven development framework. Four stages—context engineering, prompt iteration, offline evaluation, and production deployment—are connected by a feedback loop where online metrics inform subsequent iterations. The correlation between offline evaluation and production outcomes validates the pipeline's predictive power.}
  \label{fig:framework-overview}
\end{figure*}

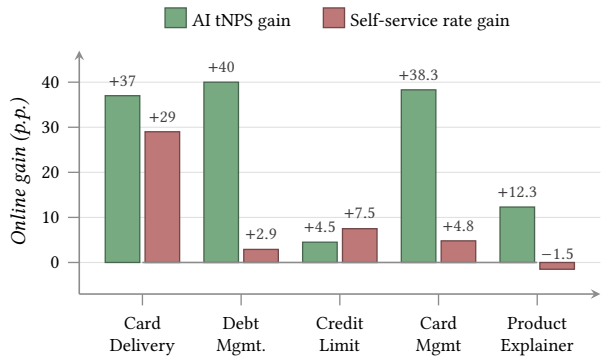
\begin{figure}[t]
    \centering
    \begin{tikzpicture}
    \begin{axis}[
        width=\columnwidth,
        height=4.8cm,
        ybar=2pt,
        bar width=13pt,
        ymin=-7, ymax=47,
        ytick={0,10,20,30,40},
        ylabel={Online gain (p.p.)},
        ylabel style={font=\small\itshape, yshift=-2pt},
        symbolic x coords={cd, debt, credit, cm, pe},
        xtick=data,
        xticklabels={{Card\\Delivery}, {Debt\\Mgmt.}, {Credit\\Limit},
                     {Card\\Mgmt}, {Product\\Explainer}},
        x tick label style={align=center, font=\footnotesize, yshift=-1pt},
        y tick label style={font=\footnotesize},
        ymajorgrids=true,
        major grid style={line width=0.4pt, draw=black!12},
        axis lines=left,
        axis line style={black!45, line width=0.6pt},
        tick style={black!45, line width=0.5pt},
        tick align=outside,
        axis on top=false,
        enlarge x limits=0.16,
        clip=false,
        legend style={font=\footnotesize, at={(0.5,1.04)}, anchor=south,
                      legend columns=2, draw=none, fill=none,
                      /tikz/every even column/.append style={column sep=10pt}},
        legend image code/.code={\draw[#1] (0cm,-0.08cm) rectangle (0.28cm,0.18cm);},
        nodes near coords,
        nodes near coords style={font=\scriptsize\bfseries, color=black!75,
                                 /pgf/number format/fixed,
                                 /pgf/number format/fixed zerofill=false,
                                 /pgf/number format/print sign=true},
        every node near coord/.append style={anchor=south, yshift=0.5pt},
    ]
    \addplot[fill=MutedGreen, draw=MutedGreen!60!black, line width=0.5pt]
        coordinates {(cd,37) (debt,40) (credit,4.5) (cm,38.3) (pe,12.3)};
    \addplot[fill=MutedRed, draw=MutedRed!60!black, line width=0.5pt]
        coordinates {(cd,29) (debt,2.9) (credit,7.5) (cm,4.8) (pe,-1.5)};
    \draw[black!45, line width=0.6pt] (axis cs:cd,0) -- (axis cs:pe,0);
    \legend{AI tNPS gain, Self-service rate gain}
    \end{axis}
    \end{tikzpicture}
    \caption{Online A/B-test gains for five production agents built with the same eval-driven framework. AI transactional NPS (tNPS) improves across all five domains; self-service rate improves in four (Product Explainer's slight dip stems from transient LLM request failures). On most use cases the resulting AI tNPS lands within a few percentage points of expert human agents (Table~\ref{tab:use-case-summary}).}
    \label{fig:headline-results}
\end{figure}

While AI agents have seen rapid success in coding and deep research~\cite{cursor2024,openai2025deepresearch,google2025antigravity}, their potential in many other domains remains largely untapped. In \textbf{customer support (CS)}, they could transform the way organizations handle millions of interactions — evolving from scripted chatbots to systems that personalize experiences, maintain rich context, take actions through tools, and resolve complex issues end-to-end. Crucially, they can also act as co-pilots for human agents, handling routine cases and enabling people to focus on the hardest, long-tail problems.

\subsection{Unique challenge of building CS AI agents} 

Compared to general-purpose AI agents, CS agents impose distinct requirements:

\begin{itemize}[leftmargin=1em]
    \item \textbf{Higher quality bar.} In CS, failure is costly: a single poor interaction can erode trust and drive churn. This creates a stricter reliability threshold than in many research or coding settings. In a banking environment, any mishap in actions can lead to significant amounts of losses in a single case.
    \item \textbf{Sensitive data handling.} CS agents operate over real customer data (e.g., balances, transactions, addresses), which introduces privacy, security, and regulatory constraints.
    \item \textbf{Narrower scope, deeper specialization.} CS agents typically use a small, domain-specific toolset (often 5--15 tools), but must handle substantial edge-case diversity within that scope.
    \item \textbf{Graceful handoff to humans.} CS agents work within a human-in-the-loop workflow. When automation falls short, they must transfer control with full context preserved, since customer satisfaction reflects the end-to-end journey—not just the automated segment (for a high-level architecture, see \textbf{Figure~\ref{fig:ai-agent}}).
\end{itemize}

\subsection{Our Contributions}

\begin{figure}
    \centering
    \includegraphics[width=1\linewidth]{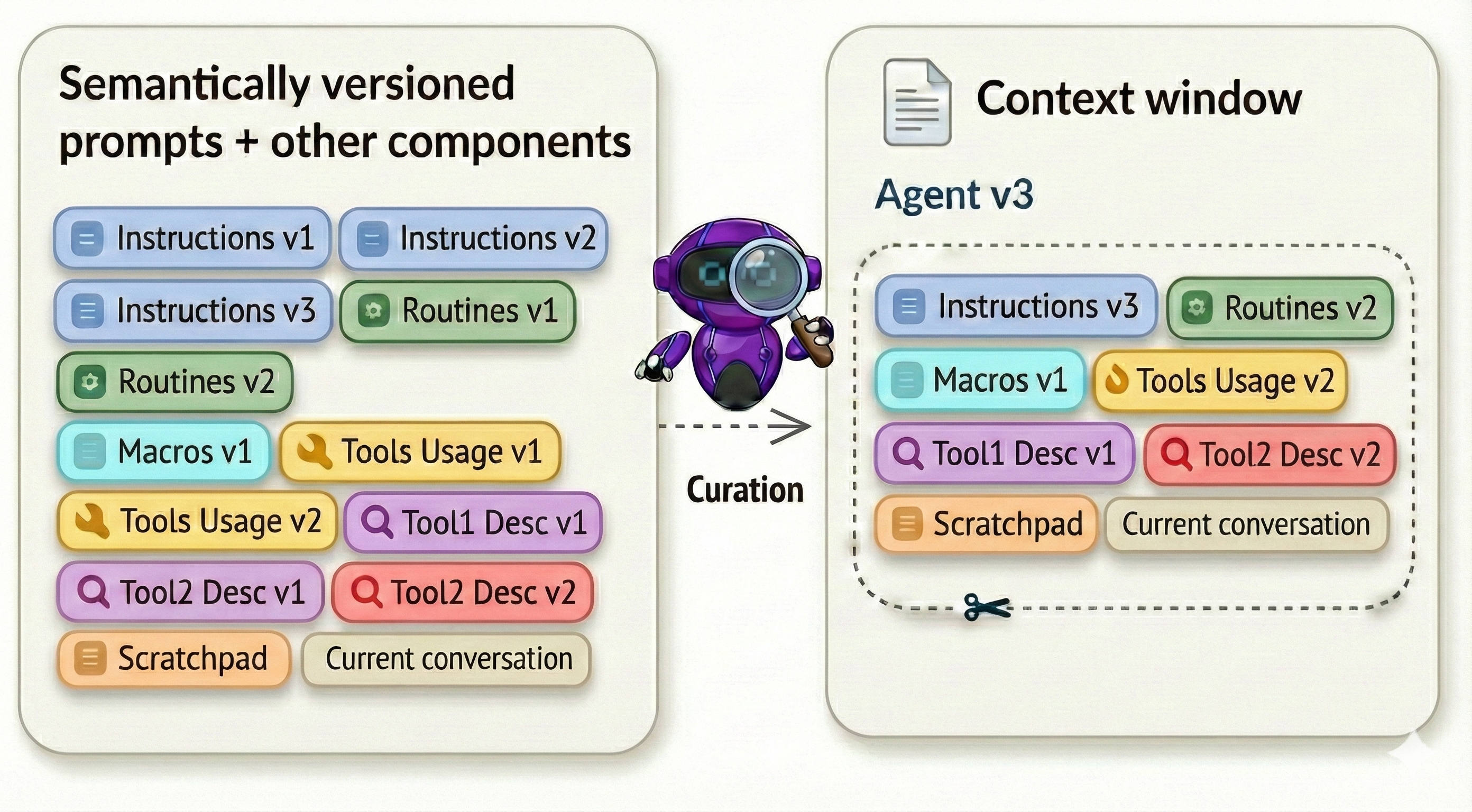}
    \caption{Our CS Agent context engineering is modular, and semantically versioned. Independently versioned components can be composed into distinct agent configurations.}
    \label{fig:context-eng7-4}
\end{figure}

We present a highly-scalable, production-tested design for building CS AI agents. Our design is based on evaluation and measurement, working backward from delivering a great customer experience (\textbf{Figure \ref{fig:framework-overview}}). We validate our design on real-world CS use cases for a company that serves 100M+ users. Our contributions can be summarized as follows:

\begin{itemize}[leftmargin=1em]

\item \textbf{A unified, evaluation-driven framework} for developing 
production CS agents, connecting agent architecture, prompt versioning, 
offline evaluation, and online measurement in a closed loop 
(\textbf{Figure~\ref{fig:framework-overview}}). A central insight is that 
evaluation pipeline quality directly determines iteration velocity.

\item \textbf{Modular context engineering.} We propose a prompt 
architecture in which each component---instructions, routines, macros, 
tool descriptions, and working memory---is independently versioned, 
enabling controlled evolution and reuse (\textbf{Figure~\ref{fig:context-eng7-4}}).

\item \textbf{A rigorous evaluation pipeline.} We introduce a 
calibrated LLM-as-Judge pipeline with (a) inter-rater reliability 
requirements enforcing minimum human-agreement thresholds before 
trusting automated judgments, (b) GEPA-optimized~\cite{gepa2025} judge prompts that 
align with human criteria, and (c) cross-model validation showing 
that well-specified rubrics yield stable scores across model families, 
sizes, and vendors.

\item \textbf{Prod-validated results at scale across five use cases.} We report A/B-tested deployments of five production agents built with the same framework, spanning card delivery, debt management, credit-limit support, card management, and product explanation (\textbf{Figure~\ref{fig:headline-results}}). For our headline card-delivery deployment, we achieve a \textbf{37~p.p.} absolute improvement in AI transactional net promoter score (tNPS - a measure of quality) and \textbf{29~p.p.} in self-service rate (SSR - a measure of automation) when compared to previous agent variants; the final tNPS is only 10~p.p. lower than expert human-agent tNPS. The four further agents show consistent customer-satisfaction gains in their respective domains (Section~\ref{sec:generalization}). Across most of these deployments the AI agent's tNPS lands within a few percentage points of expert human agents (within ${\sim}1$--$10$~p.p. for card delivery, credit-limit, card management, and product explanation), with a larger gap remaining in the hardest domain, debt management. Crucially, we demonstrate strong correlation between offline eval scores and online business metrics (\textbf{Figure~\ref{fig:offline-online}}), validating that eval-driven development reliably predicts production impact.

\end{itemize}

\vspace{-3.5mm}
\section{Related Work}

\noindent\textbf{System Architectures and Benchmarks.} Prior CS systems span modular designs - Obadinma et al.~\citep{obadinma-etal-2022-bringing}'s Neural Agent Assistant (intent classification, retrieval, generation), Airbnb's policy-to-schema assistant~\citep{su-etal-2025-llm}, and out-of-scope detection for graceful human handoff~\citep{qian-etal-2022-distinguish}. On the data side, NatCS~\citep{gung-etal-2023-natcs} offers a multi-domain support corpus, CSConv/RoleCS~\citep{zhu-etal-2026-evaluating} a staged conversation framework with synthetic training data, and Mendon\c{c}a et al.~\citep{mendonca-etal-2023-dialogue} a bilingual corpus annotated for emotion and dialogue quality.

\noindent\textbf{Feedback, Trust, and Evaluation.} Feng et al.~\citep{feng-etal-2023-schema} model user satisfaction via goal-tracking task slots; F{\o}lstad et al.~\citep{folstad-etal-2024-breakdown} show early breakdowns erode trust; and Reinhard et al.~\citep{reinhard-etal-2024-generative} frame generative AI as augmenting human agents. Mohammadi et al.~\citep{mohammadi-etal-2025-survey} survey LLM-agent evaluation by objective and mode (offline vs.\ online), noting enterprise challenges such as policy compliance and long-horizon reliability.
\section{Background}

\subsection{AI Agents}

\begin{figure}
  \centering
  \includegraphics[width=\columnwidth]{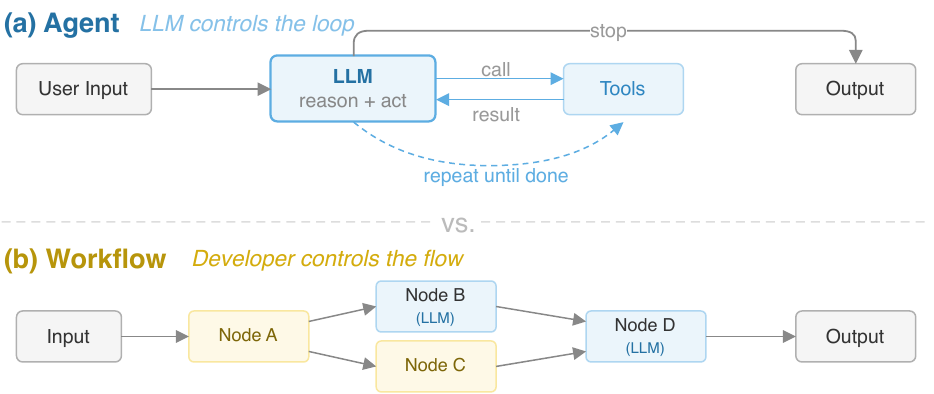}
  \caption{(a) Our CS agents follow the ReACT pattern: the LLM iteratively calls tools and reasons over results. (b) A workflow is a predefined DAG where the developer controls the execution flow; individual nodes may or may not use an LLM.}
  \label{fig:agent-vs-workflow}
\end{figure}

While there is no consensus on the definition of \textit{AI Agents}, we use it to specifically refer to the ReACT paradigm~\cite{yao2023react}. ReACT stands for Reason + Act, where an LLM interleaves chain-of-thought reasoning~\cite{wei2022chain} with tool use and can autonomously work on a task (given user input) until it either completes the task, fails, or seeks further user input. In contrast, we define \textit{workflows} as predefined computation graphs or DAGs where the developer - and not the LLM - controls the execution flow. Each node in the graph may or may not invoke an LLM. \textbf{Figure~\ref{fig:agent-vs-workflow}} disambiguates between agents and workflows.

Frameworks like LangGraph~\cite{langgraph2024} blur the boundaries between workflows and agents by allowing users to build graphs where some nodes are deterministic (workflow-like) and others involve autonomous decision-making (agent-like). In practice, however, our implementations adhere closely to the ReACT framework, with targeted modifications — such as guardrails and mandatory compliance checks — to address business and regulatory requirements.


\subsection{Prompt Engineering for LLMs}

LLMs generalize to new tasks when guided by well-crafted prompts: \emph{in-context learning} supplies a few input--output examples without fine-tuning~\cite{brown2020language}, and \emph{chain-of-thought} prompting adds intermediate reasoning steps for complex tasks~\cite{wei2022chain,kojima2022large}. \emph{Modular prompting} decomposes tasks into coordinated stages---e.g., ReAct interleaves reasoning with tool-invoking actions~\cite{yao2023react}.

\subsection{Automated Prompt Optimization}
Manual prompt engineering is labor-intensive and brittle, as small wording changes can substantially alter model behavior. \emph{Automated prompt optimization} treats prompt text as an optimizable parameter, using search algorithms to maximize a target metric.
Several methods cast this as LLM-driven search. APE~\cite{zhou2022large} uses an LLM to generate and score candidate instructions. OPRO~\cite{yang2023large} iteratively proposes improved prompts conditioned on prior scores. ProTeGi~\cite{pryzant2023automatic} introduces natural-language ``gradients'' - critiques of failures - to guide beam-search refinement, while PromptBreeder~\cite{fernando2023promptbreeder} co-evolves task prompts and mutation operators.
DSPy~\cite{khattab2024dspy} provides a programmatic abstraction where LLM pipelines are defined as modular programs with typed signatures and compiled into optimized prompts. Within DSPy, GEPA~\cite{gepa2025} implements a reflection-based loop: a strong model critiques errors, synthesizes revisions, and selects candidates via Pareto-based trade-off analysis, optionally incorporating free-text human feedback.

\subsection{Evaluation}
\label{subsec:evaluation}
Evaluating the quality of LLM-generated text is a prerequisite for any iterative improvement: without reliable evaluation, teams cannot distinguish signal from noise when comparing agents .

\subsubsection{LLM-as-a-judge}

Traditional evaluation of natural language generation often relied on reference-based metrics such as {BLEU}~\cite{papineni-etal-2002-bleu} and {ROUGE}~\cite{lin-2004-rouge}, which approximate quality via surface-level overlap with references. However, for open-ended, multi-turn dialogue, these metrics correlate weakly with human judgment~\cite{liu-etal-2016-evaluate}. The \emph{LLM-as-a-judge} paradigm instead uses a strong LLM to score or rank outputs against an explicit rubric~\cite{zheng2023judge}. Given a rubric, the judge model assesses whether a response meets quality dimensions such as correctness, helpfulness, safety, or adherence to instructions. This approach scales to thousands of examples, accommodates natural language variation, and can evaluate nuanced, domain-specific criteria that resist rule-based checking.

However, LLM judges exhibit systematic biases, including position and verbosity bias~\cite{zheng2023judge}, self-enhancement bias~\cite{panickssery2024selfrecognition}, and sensitivity to rubric wording~\cite{li-etal-2025-generation,arabzadeh2025prompt_sensitivity_relevance_judgment,yamauchi2025design_choices_reliability}. These failure modes motivate two strategies. First, judge ensembles — using multiple models or prompts — can mitigate individual biases, analogous to multi-annotator schemes~\cite{verga2024judgesjuries}. Second, automated prompt optimization for evaluation criteria can systematically search for rubric formulations that maximize agreement with human annotations, reducing the brittleness of manually authored prompts.

\subsubsection{Measuring reliability of judges}
Whether evaluation is performed by humans, LLMs, or both, the fundamental question is how much we can trust the ratings. \textbf{Inter-rater reliability (IRR)} quantifies agreement among independent raters beyond chance. Cohen's $\kappa$~\cite{cohen1960coefficient} assesses pairwise agreement on categorical labels:
\begin{equation}
\kappa = \frac{p_o - p_e}{1 - p_e},
\end{equation}
where $p_o$ is observed agreement and $p_e$ is expected agreement under chance. Fleiss's $\kappa$ generalizes this to any number of raters~\cite{fleiss1971measuring}, and Krippendorff's $\alpha$ extends it to various data types and missing data~\cite{krippendorff2004reliability}. High agreement (e.g., $\kappa$ or $\alpha$ exceeding $0.8$) instills confidence that quality labels are well-defined and consistently applied.
Cohen's $\kappa$ naturally extends to the LLM-as-a-judge setting, where two judge models take the role of two annotators. By comparing $\kappa$ across judge pairs before and after prompt optimization, we evaluate whether GEPA improves rubric stability across model architectures.

\subsubsection{Online Evaluation}

Customer-support agents ultimately succeed or fail with real users, so online metrics from deployment are central to eval-driven development. Many teams track \textbf{CSAT (Customer Satisfaction)}, typically collected via post-interaction surveys on a Likert scale (e.g., ``Were you satisfied with the help you received?'')~\cite{likert1932technique}. Another widely used measure is \textbf{Net Promoter Score (NPS)}, which estimates how likely a user is to recommend the service~\cite{reichheld2003one}. For CS, organizations often use \textbf{Transactional NPS (tNPS)} - an NPS variant asked immediately after a support interaction. It is important to note we measure tNPS for AI Agents and human experts separately, using the support funnel described in Figure~\ref{fig:ai-agent}.

Beyond satisfaction, teams track effectiveness and deflection. \textbf{Self-service rate (SSR)} measures the fraction of support needs resolved via automated or self-serve channels without human escalation. High SSR and task success, together with strong CSAT and tNPS, indicate the agent is delivering effective support. It is worth calling out that there is usually a trade-off between SSR and tNPS, as you can typically improve overall tNPS by sacrificing SSR. One example is directly transfering hard cases or frustrated users to humans and only keeping the easier cases to be solved by the AI agent. 

\section{System Design}

\subsection{Design Principles}

We propose the following design principles for CS AI agents:

\begin{itemize}[leftmargin=1em]
\item \textbf{Reliable.} Consistent behavior across interactions with graceful fallback to human agents, preserving full conversational context.
\item \textbf{Correct.} Responses grounded in verified data from tools and knowledge bases — no hallucinated details or incorrect promises.
\item \textbf{Delightful.} Empathetic, concise, and natural interactions reflecting the organization's brand.
\item \textbf{Compliant.} Adherence to regulations, data-handling policies, and internal business rules.
\item \textbf{Measurable.} Quality quantifiable through offline evaluations and online business metrics (tNPS, SSR, task success rate).
\end{itemize}

\subsection{Context Engineering for Agents}
\label{subsec:context-engineering}


\begin{figure}
  \centering
  \includegraphics[width=0.9\columnwidth]{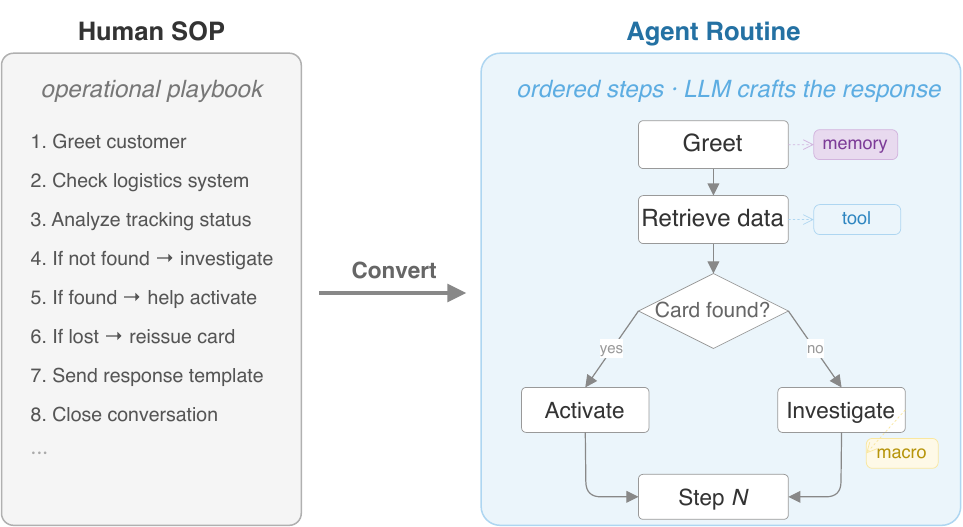}
  \caption{An example of a routine - translation of a human standard operating procedure (SOP) into ordered, agent-executable steps.}
  \label{fig:routine}
\end{figure}


Prompt engineering—crafting textual instructions that steer LLM behavior—understates the scope of what a production system actually needs. For a CS agent, instructions are only one component occupying the context window. We adopt the framing of \textbf{context engineering}: the systematic design and assembly of the entire input presented to the model at every turn. \textbf{Figure~\ref{fig:context-eng7-4}} shows our context engineering design:

\begin{itemize}[leftmargin=1em]
\item \textbf{Instructions.} The agent's role, definition, behavioral constraints, compliance guardrails, communication style, output specifications, and response construction guidelines. Instructions encode \emph{how} the agent should behave.
\item \textbf{Routines.} Most customer support units within organizations issue standard operating procedures (SOPs) for human agents to assist customers using an automated process. We convert these SOPs into routines - ordered, step-by-step procedures that guide the agent through a domain-specific conversation flow (see \textbf{Figure~\ref{fig:routine}}). For example, a card delivery investigation routine might prescribe: (1) greet and frame the case, (2) retrieve logistics data, (3) run address-type-specific investigation questions, (4) offer reissue if unresolved.
\item \textbf{Macros.} Pre-written response templates for common scenarios — serve as modular building blocks that the agent adapts and condenses rather than composing responses from scratch while still allowing natural variation in phrasing.
\item \textbf{Tool specifications and usage.} Each tool is described independently — name, input schema, expected output, and invocation conditions — so tool sets can be recombined across configurations. Orchestration rules then govern how tools are called in concert: parallel invocation strategies, transient error handling, confirmation gating (only confirm outcomes on explicit success), and grounding rules.
\item \textbf{Working memory (scratchpad).} A persistent state object that tracks information gathered across the conversation: collected parameters, prior tool outputs, current routine step. The scratchpad prevents redundant tool calls while also enabling long-term personalization.
\end{itemize}

Each component is independently versioned, allowing a new agent configuration to be composed by selecting specific versions of each module (\textbf{Figure~\ref{fig:context-eng7-4}}). This transforms prompt iteration from monolithic rewrites into targeted, testable changes.

\subsection{Infrastructure and Tool Design}
\label{sec:tool-design}

A common failure mode in agent development is beginning with prompts and adding infrastructure later. Our experience indicates the opposite ordering is more effective: the data and action layers should be established first, because they constitute the durable foundation that persists across model generations.

\subsubsection{Infrastructure layers}

Our agent system rests on three infrastructure layers:

\begin{itemize}[leftmargin=1em]
\item \textbf{Identity and authentication.} Secure, consent-aware access to customer data. 
\item \textbf{Action APIs.} Idempotent, transactional endpoints that execute real business logic (e.g., reissuing a card, adjusting a payment plan). Idempotency is critical: because LLMs may retry tool calls on transient failures or ambiguous results.
\item \textbf{Audit trails.} Comprehensive logging of every tool call—inputs, outputs, latency, and errors—with masking of sensitive fields, strict access controls, and retention schedules aligned with legal and regulatory requirements—for compliance, debugging, and post-hoc analysis.
\end{itemize}

\subsubsection{Tool design principles}

Tool design directly affects agent reliability:

\paragraph{Compose deterministic flows in code, not in prompts.} When a task requires chaining multiple data sources—for example, checking carrier tracking status \emph{and} querying the internal order management system—we wrap the orchestration into a single composite tool rather than relying on the LLM to sequence two separate calls. This reduces latency, eliminates sequential-dependency errors where the model calls tools in the wrong order or drops an intermediate result, and reserves the LLM's reasoning capacity for tasks that genuinely require judgment.

\paragraph{Tool descriptions are prompts.} The schema and docstring of each tool function as part of the agent's prompt context (\S\ref{subsec:context-engineering}). Ambiguous or incomplete tool descriptions lead to misuse—incorrect parameters, calls at the wrong time, or misinterpretation of outputs. We treat tool documentation with the same rigor as other prompt components: each tool includes a natural-language description, a typed input schema, the expected output format, example invocations, and explicit guidance on when \emph{not} to call it. 

\paragraph{Minimize output surface area.} Tools return only the fields the LLM needs to reason about, not the full upstream API response. Raw responses often contain dozens of internal fields that consume context-window budget and increase the risk of the model latching onto irrelevant details.

\paragraph{Design for idempotency.} Because LLMs may retry tool calls on transient failures or ambiguous results, every action tool must be safely re-callable without side effects.

\subsection{Development Lifecycle}
\label{subsec:dev-lifecycle}

\textbf{Figure~\ref{fig:framework-overview}} depicts our development process using two nested iteration loops.

The pipeline flows through four stages. First, \emph{Agent Architecture} (\textcircled{\small 1}) defines the modular prompt structure that constitutes the agent's context (\S\ref{subsec:context-engineering}). These components are passed to \emph{Prompt Versioning} (\textcircled{\small 2}), where domain experts iterate on individual modules, producing a sequence of candidate versions ($v_1 \rightarrow v_2 \rightarrow \ldots \rightarrow v_N$). Each candidate is then evaluated in \emph{Offline Evaluation} (\textcircled{\small 3}) using real and/or simulated conversations scored by calibrated LLM judges. Only versions that pass offline evaluation—demonstrating improved or stable metrics relative to the current baseline—are promoted to \emph{Production} (\textcircled{\small 4}), where they are deployed via A/B tests and measured on tNPS, SSR, and other business metrics.

Two feedback loops drive continuous improvement:

\begin{itemize}[leftmargin=1em]
\item \textbf{Fast loop } Offline evaluation results feed directly back into prompt versioning. When an evaluator flags a failure—for example, the agent omitting address verification—the team traces the failure to the responsible prompt component, applies a targeted fix, and re-evaluates. Fixes are surgical: modifying one routine step does not require re-testing the entire prompt.
\item \textbf{Slow loop} Production metrics reveal patterns that offline evaluation may not capture—for instance, a consistent tNPS gap on a specific customer segment, or a rising escalation rate on a newly launched product. These signals inform architectural changes: adding new routines, redesigning tool interfaces, or restructuring the context allocation.
\end{itemize}

This two-loop structure helps prove that evaluation quality determines iteration velocity. A reliable fast loop—enabled by calibrated judges and rigorous inter-rater agreement (\S\ref{subsec:evaluation})—allows teams to iterate confidently without waiting for slow, sparse online signals.

\subsection{Deployment Strategy}
\label{sec:deployment}

Deploying an agent to production at scale requires a progressive rollout strategy that balances iteration velocity with risk management.

\subsubsection{Progressive rollout}
New agent versions are deployed at a small percentage (1--5\%) while monitoring key metrics (SSR, tNPS, escalation rate, error rate).

\subsubsection{Confidence-based escalation}
The agent operates within a human-in-the-loop architecture (Figure~\ref{fig:ai-agent}). When the agent encounters situations outside its defined routines, fails to retrieve the necessary data, or detects low confidence in its proposed response, it transfers the conversation to a human agent.

\subsubsection{Model portability}
Because foundation models evolve rapidly, our system is designed to accommodate model transitions with minimal disruption. The modular context engineering architecture (\S\ref{subsec:context-engineering}) enables targeted prompt adjustments. We maintain per-model evaluation baselines and run the full evaluation suite before any model migration, treating model upgrades as first-class prompt changes subject to the same eval-driven process described in subsequent sections. More details using a case study in Section~\ref{section:methods}

\section{Building a CS Agent at Scale - A Case Study}
\label{section:methods}

We validate our framework on a \textbf{card delivery} agent — one of the highest-volume intents at our organization, handling ``Where is my card?'' inquiries. It exercises the full framework: it requires multi-step reasoning over real-time logistics data, context-dependent branching (e.g., address type determines the investigation path), and careful grounding to avoid misstating delivery timelines.

\subsection{Agent Overview}

The card delivery agent's mission is to resolve card status inquiries. A customer contacts support after ordering a new card (due to first issuance, loss, theft, damage, or expiration), and the agent must determine the card's current status and guide the customer to resolution.

\paragraph{Routine.} The agent follows a routine (see Figure~\ref{fig:routine} for the general pattern) with the following steps: (1) greet and frame the case, (2) retrieve logistics and tracking data via tools, (3) analyze delivery status against expected timelines, (4) run address-type-specific investigation questions (e.g., apartment access, concierge, PO box), and (5) offer card reissue if the issue cannot be resolved.

\paragraph{Tools.} The agent has access to a small, domain-specific toolset:
\begin{itemize}[leftmargin=1em]
\item \texttt{get\_customer\_profile} — retrieves identity, address, and account details for the authenticated customer.
\item \texttt{check\_card\_delivery} — queriesthe order management system for tracking status, expected delivery date, and delivery confirmation.
\item \texttt{reissue\_card} — initiates a card reissue to the same or a new address.
\end{itemize}

\subsection{Building Evals}

We now discuss how we built \textit{LLM-as-a-judge} evaluators for this use case. Using rigorous evals, we A/B tested 11 variants of the agent online, leading to large gains (see Results for details).

\noindent\textbf{Data Collection and Annotation.} 
To rigorously evaluate the performance of the agent scenarios, we constructed a dataset spanning five evaluation categories (Evals E1--E5), see Table~\ref{tab:evals-dataset-description} for details. This and all other datasets used in this work are privacy-preserving datasets derived from real customer-support conversations through the data-minimization and pseudonymization pipeline described in Section~\ref{sec:data-protection}. Ground-truth binary labels were obtained through a multi-annotator procedure involving three operations analysts with specialized domain knowledge of the target business logic. For each sample, analysts provided both a classification label and a detailed textual explanation justifying their decision. The final ground truth for each sample was determined by majority vote among the three annotators. The accompanying textual explanations were retained as semantic feedback signals for the subsequent optimization phase. \textbf{It is important to note that only a few hundred high quality binary labels are enough to build an effective eval.}

\begin{table}[!htb]
\centering
\resizebox{\columnwidth}{!}{%
\begin{tabular}{lccc}
\toprule
\textbf{Description} & \textbf{\# Count} & \textbf{\# Class} & \shortstack{\textbf{Freq.} \\ \textbf{Min. Class}} \\
\midrule
E1: Agent Reissue Failure & 888 & 2 & 0.49 \\
E2: Customer Input Verification & 90 & 2 & 0.22 \\
E3: Card Delivery Data Check & 816 & 2 & 0.43 \\
E4: Response Conciseness & 246 & 2 & 0.49 \\
E5: Resolution Completeness & 90 & 2 & 0.49 \\
\bottomrule
\end{tabular}%
}
\caption{Summary statistics of the evaluation datasets.}
\label{tab:evals-dataset-description}
\end{table}

\noindent\textbf{Prompt Optimization.}
We used GEPA~\cite{gepa2025} within the DSPy framework to automatically refine the eval instructions. 

All classification evaluations used GPT-4.1-mini as the base model and GPT-5.1 as a reflection model to critique errors, synthesize revisions, and propose semantically grounded prompt edits informed by analyst feedback. We ran GEPA with \texttt{auto=light} (approximately 500 iterations), reflection minibatch size $=3$, and Pareto-based candidate selection to balance trade-offs across optimization criteria. We also provided the optimizer with the free-text rationales collected during annotation, enabling the reflection model to anchor prompt updates in expert reasoning rather than surface heuristics.

We compared against (i) a \emph{majority-class} baseline that always predicts the most frequent label, and (ii) a \emph{starter-prompt} baseline consisting of a brief, manually written instruction set. We repeated each experiment five times and report mean accuracy and 95\% confidence interval across runs.

We stratified the annotated dataset into train, validation, and test splits using a 40\%/30\%/30\% ratio. During the optimization phase, GEPA utilized the train and validation set to iteratively refine prompt candidates based on the provided feedback. Final performance results reported were measured on the held-out test set.

\noindent\textbf{Inter-rater Agreement Analysis.}
To validate the robustness of evals after optimization, we analyzed the consistency of predictions across several state-of-the-art language models: GPT-4o, GPT-4o-mini, GPT-4.1, GPT-4.1-mini, GPT-5, o3, o3-mini, all operated under data-processing agreements in controlled environments. The model temperature was set to 0, and all other hyperparameters were left at their default values. This analysis was conducted in two phases: first, using the manual starter prompt and subsequently using the optimized prompt.

We quantified agreement between model pairs using Cohen's kappa ($\kappa$) as described in the background section. By comparing $\kappa$ before and after prompt optimization, we evaluate whether GEPA improves the stability of the rubric such that it induces more consistent judgments across different model architectures.

\begin{figure*}[t]
\centering

\begin{subfigure}[t]{0.45\linewidth}
  \centering
  \includegraphics[width=\linewidth]{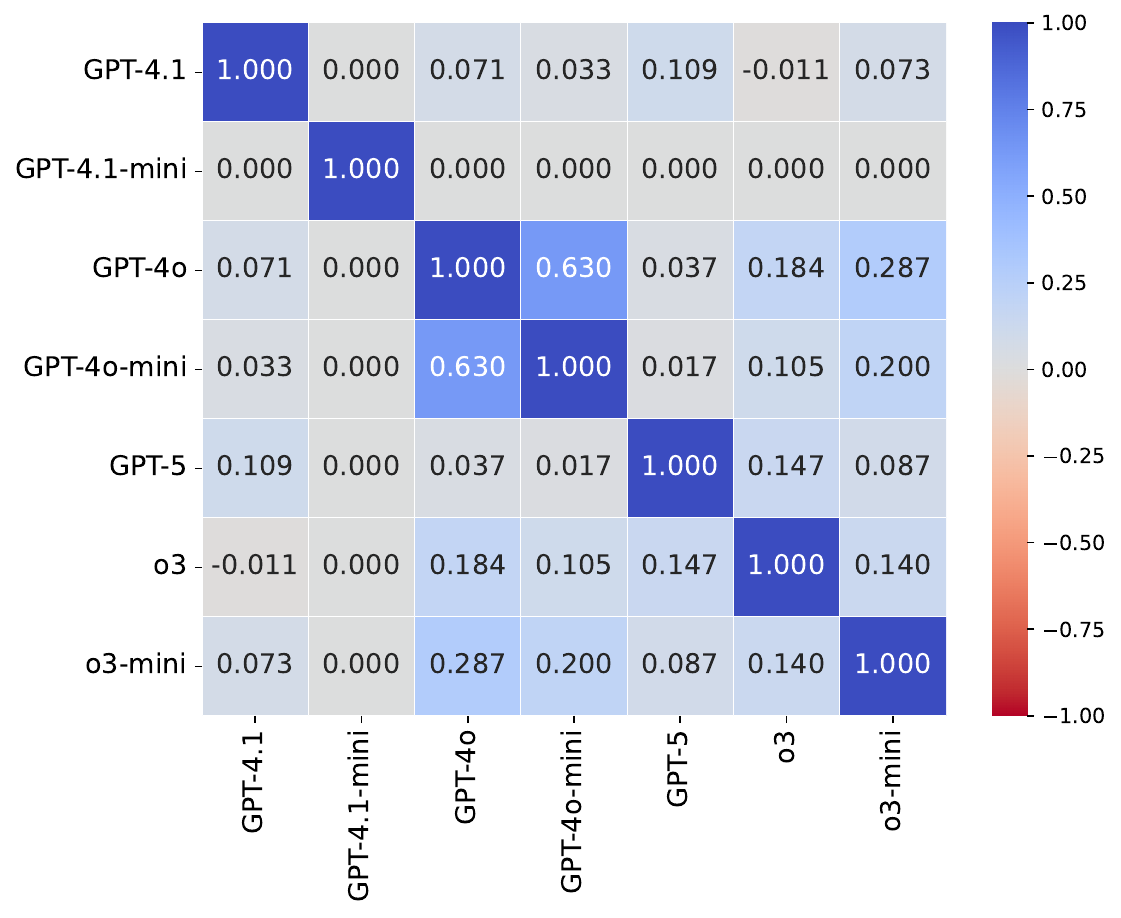}
  \caption{Starter prompt}
  \label{fig:kappa_before}
\end{subfigure}
\hfill
\begin{subfigure}[t]{0.45\linewidth}
  \centering
  \includegraphics[width=\linewidth]{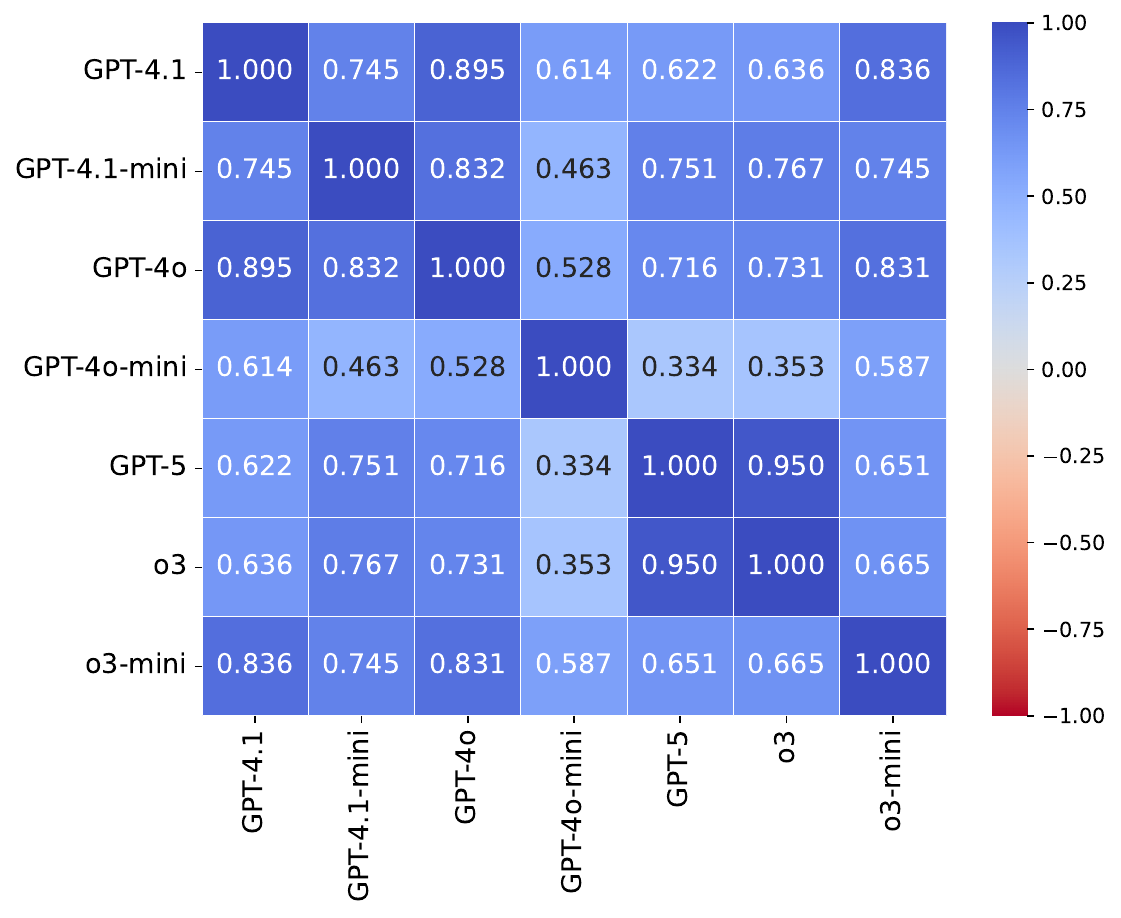}
  \caption{GEPA-optimized prompt}
  \label{fig:kappa_after}
\end{subfigure}

\caption{Eval E1 - Pairwise Cohen's $\kappa$ heatmaps among seven language models, comparing the starter prompt (left) and the GEPA-optimized prompt (right).}
\label{fig:kappa_pairwise}
\end{figure*}

\section{Offline and Online Results}

\subsection{Prompt Optimization}

\begin{table}[htb]
    \centering

    \begin{tabular}{lccc}
        \toprule
        \multirow{2}{*}{\textbf{Eval}} & \multicolumn{3}{c}{\textbf{Accuracy (\%)}} \\
        \cmidrule(lr){2-4}
         & \textbf{Baseline} & \textbf{Starter Prompt} & \textbf{Optimized Prompt} \\
        \midrule
        E1 & 51.11 & $77.78_{\pm 0.49}$ & $\mathbf{82.00}_{\pm 0.09}$ \\
        E2 & 77.78 & $68.88_{\pm 3.41}$ & $\mathbf{88.89}_{\pm 0.00}$ \\
        E3 & 56.62 & $58.07_{\pm 0.26}$ & $\mathbf{73.01}_{\pm 0.76}$ \\
        E4 & 51.11 & $74.00_{\pm 0.95}$ & $\mathbf{76.89}_{\pm 0.79}$ \\
        \bottomrule
    \end{tabular}
    \caption{Evaluator accuracy (mean with 95\% confidence intervals in subscript) before and after prompt optimization.}
    \label{tab:evaluator_accuracy}

\end{table}

Table~\ref{tab:evaluator_accuracy} summarizes the performance and demonstrates that automated optimization consistently outperforms both the majority-class baseline and manual prompt engineering across all tasks. Notably, in E3, the optimization delivered a substantial improvement, raising accuracy from a manual baseline of $58.07$ to $73.01$. This illustrates the framework's capability to capture complex evaluation criteria where manual prompting failed to align with expert consensus. Furthermore, the narrow confidence intervals observed across all optimized prompts (ranging from $0.00$ to $0.79$) indicate that the GEPA optimization process yields robust prompts that are consistent across multiple inference runs. \textbf{A sample before-and-after prompt for Eval E2 can be found in Appendix~\ref{app:prompt-evolution}}.

A critical insight emerged from E2, a binary evaluator characterized by a high class imbalance (baseline accuracy of $77.78$). In this instance, the manual starter prompt degraded performance to $68.88$, performing worse than the majority-class baseline. \textbf{This highlights the risks inherent in manual prompt engineering, where human-written instructions may inadvertently introduce noise or bias that contradicts the underlying task distribution}. Conversely, optimization successfully recovered performance, achieving an accuracy of $88.89$ by aligning the GPT-4.1-mini base model with the rigorous logic of the human specialists.

For tasks with more balanced class distributions, specifically Eval E1 and E4, where baselines were approximately $51.11$, the Optimized Prompt demonstrated high reliability, achieving $82.00$ and $76.89$ accuracy, respectively. However, the positive lift from the starter prompt was small compared to E2.

\subsection{Inter-rater Agreement}

Figures~\ref{fig:kappa_before} and ~\ref{fig:kappa_after} present the pairwise Cohen's Kappa coefficients for Eval E1, calculated before and after GEPA prompt optimization, respectively.

The starter prompt resulted in negligible agreement across the majority of model pairs. For instance, the interaction between the GPT-4.1 and GPT-5 families yielded a low coefficient ($\kappa=0.109$). Notably, the agreement between GPT-4.1 and GPT-4.1-mini was $0.00$. Inspection of the predictions revealed that the unoptimized GPT-4.1-mini assigned a single class to all test instances, resulting in zero variance and a consequent $\kappa$ of zero. This lack of consensus suggests that the manual starter prompts were ambiguous, leading the models to rely on internal priors.

Conversely, the GEPA-optimized prompts led to a substantial convergence in judge behavior. Post-optimization, the agreement between GPT-4.1 and GPT-4.1-mini increased to $\kappa=0.745$, while the agreement between GPT-4.1 and GPT-4o reached $\kappa=0.895$. The highest consensus was observed between the reasoning-intensive models, with GPT-5 and o3 achieving near-perfect alignment $\kappa=0.950$. These results indicate that the optimized prompts effectively grounded the evaluation criteria, rendering the judging process largely invariant to the specific model architecture.

\subsection{Online Experiments}

We used the aforementioned evals to deploy 10 different variants of the card-delivery use case agent and compared them against our legacy customer support solution across a large number of real customer interactions.

Each variant was launched at 1\% traffic, and subsequently increased to larger percentages. The evaluator scores were used to exclude non-promising variants. The baseline variant (V0) was limited to a knowledge-base retrieval tool and a human-transfer tool, unable to perform domain actions. Subsequent versions progressively added: a structured routine derived from SOPs; domain-specific tools for logistics tracking and customer data retrieval; frustration detection and proactive escalation; self-contained card reissue capability, eliminating the need for human handoff on the most common resolution path; tool-selection hierarchies and conciseness budgets to curb verbosity and tool misuse — each change driven by failures surfaced through the offline evals.

\subsubsection{Business Metric Impact}

The AI agent delivered a \textbf{37} percentage point
improvement in AI tNPS, reflecting substantially higher customer satisfaction. Additionally, we observed a \textbf{29} p.p increase in SSR, indicating that more customers successfully resolved their issues without escalation to human agents. These gains translate directly to improved customer experience and reduced operational costs at scale. Notably, the AI tNPS came within \textbf{10}~p.p. of the expert human-agent tNPS score.

\definecolor{PosGreen}{RGB}{0,153,0}
\definecolor{NegRed}{RGB}{204,0,0}

\subsubsection{Model Ablation}

We evaluated the impact of the underlying base model by comparing agent variants built on GPT-4.1 versus GPT-5 (these constitute 2 of the 10 variants we tried). Notably, transitioning to GPT-5 required prompt adaptation:
GPT-5 exhibits stricter instruction-following behavior compared to GPT-4.1, which would implicitly
infer unstated requirements. For instance, GPT-5 failed to call contextual tools before responding
unless explicitly instructed, whereas GPT-4.1 inferred this requirement from context. To address this, we introduced a \emph{tool-first principle} in the Instructions component - a single directive
mandating tool calls before composing responses when customer-specific data is required. This minimal
intervention restored the desired behavior.

\begin{figure}[t]
  \centering
  \includegraphics[width=0.95\linewidth]{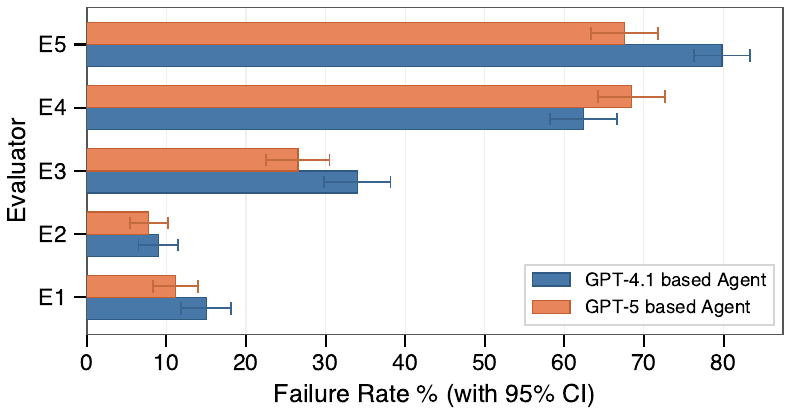}
\caption{Evaluation failure rates for GPT-4.1 and GPT-5 with 95\% confidence intervals. GPT-5 almost always achieves lower failure rates.}
  \label{fig:model-ablation}
\end{figure}

Figure~\ref{fig:model-ablation} shows the eval failure rates for both variants for each binary evaluator with 95\% confidence intervals. The GPT-5-based agent achieved reductions across all evaluation categories, except E4 (GPT5 was generating more verbose answers). \textbf{Crucially, the GPT-5 variant resulted in better online metrics, correlating with its offline eval improvements when compared to the GPT-4.1 variant}.

These results highlight a key insight: more capable models with stricter instruction-following require more explicit prompts but can deliver superior performance when properly guided. The modular prompt architecture enabled rapid, targeted modifications.

\subsubsection{Offline-Online Correlation}

\begin{figure}[!htb]
    \centering
    \includegraphics[width=\columnwidth]{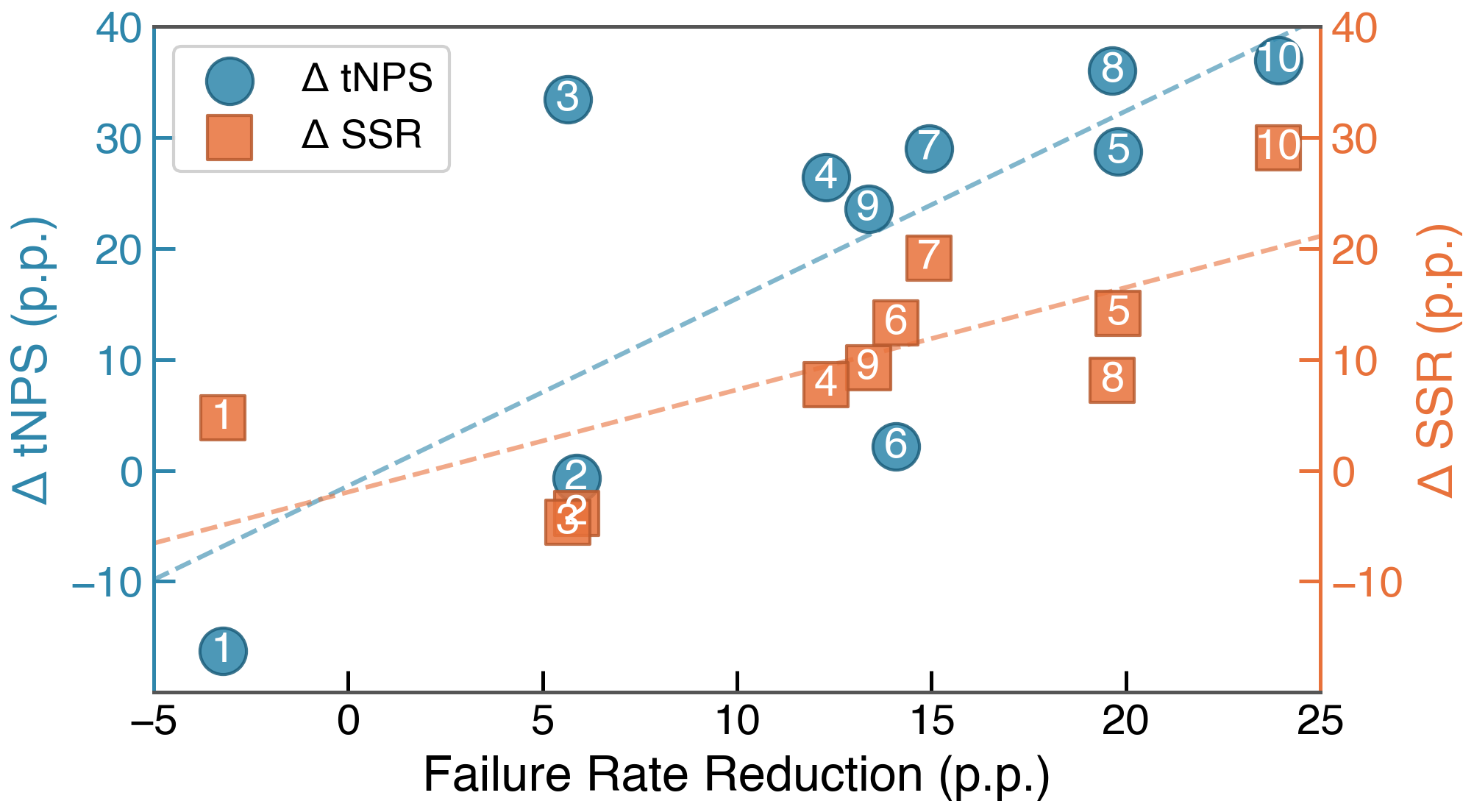}
    \caption{Offline evaluation improvements are positively correlated with online gains in both tNPS and SSR  across deployed agent variants (numbered by version), measured relative to the V0 baseline.}
    \label{fig:offline-online}
\end{figure}

A core claim of our framework is that rigorous offline evaluation enables predictive iteration:
improvements in simulation metrics should correlate with improvements in production outcomes.
Figure~\ref{fig:offline-online} validates this hypothesis by plotting the relationship between offline
evaluation performance and online tNPS across multiple agent versions (V0--V10). Each point
represents a deployed agent variant; the $x$-axis shows the change in average evaluation failure rate
relative to the baseline (V0), while the $y$-axis shows the corresponding change in tNPS.

\textbf{We observe a strong positive correlation: versions with lower eval failure rates consistently
achieve higher tNPS scores}. Notably, versions V10 and V8 achieved the largest offline improvements
(approximately 20 - 25 p.p. reductions in failure rate) and correspondingly demonstrated
the highest tNPS gains (37 p.p. relative to V0). Conversely, V6, which showed minimal offline
improvement, delivered near-zero tNPS lift.

\textbf{This correlation validates our central thesis: investment in evaluation quality - through rigorous annotation, GEPA-optimized judge prompts, and multi-model agreement analysis enables teams to iterate confidently offline}, knowing that improvements will translate to online positive gains. The framework reduces the need for expensive online experimentation, enabling fast iteration cycles.

\subsection{Generalization Across Production Use Cases}
\label{sec:generalization}

To assess whether the same eval-driven workflow generalizes to materially different customer-support domains, we report results from 4 additional agents. We describe two in detail below and summarize all five deployments in Table~\ref{tab:use-case-summary}.

\subsubsection{Debt Management Agent}

This agent helps customers regularize overdue personal-loan debt by detecting overdue status, simulating  offers across installment plans, and guiding the customer. This use case requires multi-step numerical reasoning over financial simulations and empathetic communication with customers. Using the same eval-driven workflow, the best-performing variant substantially improved customer-perceived quality -lifting AI tNPS by \textbf{40~p.p.} - while maintaining and slightly improving automation (self-service rate and return rate both moved favorably) and reducing evaluator failures by \textbf{3.5~p.p.} (Table~\ref{tab:use-case-summary}).

\subsubsection{Credit Limit Agent}

This agent handles credit-card limit questions by explaining limit breakdowns, clarifying discrepancies, and recommending alternative credit options when an increase is unavailable. Unlike card delivery, this use case requires precise grounding across multiple financial data sources and careful communication around decisions. We built three domain-specific evaluators and followed the same iteration cycle. Relative to control, the deployed agent improved AI tNPS, self-service rate, and resolution rate, and sharply cut its evaluator failure rates---for example, account-specific-context and customer-empathy failures dropped by \textbf{14.95} and \textbf{11.32~p.p.}, respectively (Table~\ref{tab:use-case-summary}).

\subsubsection{Cross-Use-Case Summary}

Table~\ref{tab:use-case-summary} summarizes the online results across the five production deployments.

\begin{table}[htb]
    \centering
    \small
    \setlength{\tabcolsep}{4pt}
    \begin{tabular}{lccc}
        \toprule
        \textbf{Use Case} & \textbf{tNPS} & \textbf{SSR} & \textbf{$\Delta$ Human} \\
                          & \textbf{Gain} & \textbf{Gain} & \textbf{tNPS} \\
        \midrule
        Card Delivery      & $+37$   & $+29$             & $-10$ \\
        Debt Management & $+40$   & $+2.9$            & $-23.6$ \\
        Credit Limit       & $+4.5$  & $+7.5$            & $-7.1$ \\
        Card Management    & $+38.3$ & $+4.8$            & $-1.1$ \\
        Product Explainer  & $+12.3$ & $-1.5^{\ddagger}$ & $-6.2$ \\
        \bottomrule
    \end{tabular}
    \caption{Online A/B-test results across five production agents. All results are in percentage points (p.p.); $\Delta$~Human~tNPS is the AI agent's tNPS minus the expert human-agent tNPS (negative $=$ below human). All five agents also showed substantial reductions in evaluator failure rates. $^{\ddagger}$The Product Explainer SSR dip reflects transient LLM request failures.}
    \label{tab:use-case-summary}
\end{table}
Taken together, these use cases span distinct domains - from logistics to debt management, credit management, card servicing, and product education - with different routines, tools, and evaluation rubrics. The same framework yields consistent customer-satisfaction (tNPS) gains across all five. The $\Delta$~Human~tNPS column further shows that, for four of the five agents, the AI reaches within ${\sim}1$--$10$~p.p. of expert human-agent tNPS, with the largest remaining gap in the hardest domain (debt management, $-23.6$~p.p.).

\subsection{Responsible Deployment: Data Protection and Privacy by Design}
\label{sec:data-protection}

Our agents operate in a highly regulated financial environment under strict privacy-by-design and privacy-by-default principles. All datasets used are derived from customer-support interactions through a data-minimization and pseudonymization pipeline that removes or masks direct identifiers and redacts sensitive financial fields. Access to conversation logs and evaluation datasets is role-based, audited, and time-limited.

When we use third-party foundation models (e.g., for judge ensembles or prompt optimization), we do so under Data Processing Agreements (DPAs) that contractually prohibit training on Nubank data, with requests restricted to minimized, privacy-preserving inputs and cross-border processing governed by an international transfer framework.

Finally, the agents make no unconstrained or irrevocable decisions about credit, debt management, or limits; they operate inside existing regulated processes subject to established policies, human oversight, and channels for contestation and review. Logging and audit trails apply the same safeguards without compromising data protection or customer trust.
\section{Conclusion}

We presented an eval-driven methodology for building high-quality customer-support agents. The approach couples a maintainable, modular prompt architecture with rigorous eval
design (including inter-rater reliability) and automated judge-prompt optimization via GEPA. In both
offline simulation and online A/B tests, this workflow enabled rapid iteration while preserving a
clear quality bar, yielding material gains in customer satisfaction and self-service over a legacy
support system. The same framework delivered consistent online customer-satisfaction gains across five deployed agents spanning logistics, debt management, credit-limit support, card management, and product explanation, suggesting that the methodology generalizes across customer-support domains. Future work will broaden use case coverage,  generalize to different open-source or proprietary models, and close the loop on fine-tuning by using the same eval-driven methodology to inform data curation and training objectives, applying strict privacy and data-minimization controls.

\bibliography{references}

@article{dubey2024llama,
  title         = {The {Llama} 3 Herd of Models},
  author        = {Dubey, Abhimanyu and Jauhri, Abhinav and Pandey, Abhinav and
                   Kadian, Abhishek and Al-Dahle, Ahmad and Letman, Aiesha and
                   Mathur, Akhil and Schelten, Alan and Yang, Amy and Fan, Angela
                   and others},
  journal       = {arXiv preprint arXiv:2407.21783},
  year          = {2024}
}

@article{comanici2025gemini,
  title={Gemini 2.5: Pushing the frontier with advanced reasoning, multimodality, long context, and next generation agentic capabilities},
  author={Comanici, Gheorghe and Bieber, Eric and Schaekermann, Mike and Pasupat, Ice and Sachdeva, Noveen and Dhillon, Inderjit and Blistein, Marcel and Ram, Ori and Zhang, Dan and Rosen, Evan and others},
  journal={arXiv preprint arXiv:2507.06261},
  year={2025}
}

@article{brown2020language,
  title={Language models are few-shot learners},
  author={Brown, Tom and Mann, Benjamin and Ryder, Nick and Subbiah, Melanie and Kaplan, Jared D and Dhariwal, Prafulla and Neelakantan, Arvind and Shyam, Pranav and Sastry, Girish and Askell, Amanda and others},
  journal={Advances in neural information processing systems},
  volume={33},
  pages={1877--1901},
  year={2020}
}

@article{guo2025deepseek,
  title={DeepSeek-R1 incentivizes reasoning in LLMs through reinforcement learning},
  author={Guo, Daya and Yang, Dejian and Zhang, Haowei and Song, Junxiao and Wang, Peiyi and Zhu, Qihao and Xu, Runxin and Zhang, Ruoyu and Ma, Shirong and Bi, Xiao and others},
  journal={Nature},
  volume={645},
  number={8081},
  pages={633--638},
  year={2025},
  publisher={Nature Publishing Group UK London}
}

@article{jiang2023mistral7b,
  title         = {Mistral 7B},
  author        = {Jiang, Albert Q. and Sablayrolles, Alexandre and Mensch, Arthur and Bamford, Chris and
                   Chaplot, Devendra Singh and de las Casas, Diego and Bressand, Florian and Lengyel, Gianna and
                   Lample, Guillaume and Saulnier, Lucile and Renard Lavaud, L{\'e}lio and Lachaux, Marie-Anne and
                   Stock, Pierre and Le Scao, Teven and Lavril, Thibaut and Wang, Thomas and Lacroix, Timoth{\'e}e and
                   El Sayed, William},
  year          = {2023},
  eprint        = {2310.06825},
  journal={arXiv e-prints},
  primaryClass  = {cs.CL},
  doi           = {10.48550/arXiv.2310.06825},
  url           = {https://arxiv.org/abs/2310.06825}
}

@misc{anthropic2024agents,
  title   = {Building effective agents},
  author  = {{Anthropic}},
  year    = {2024},
  url     = {https://www.anthropic.com/engineering/building-effective-agents},
  note    = {Published Dec 19, 2024. Accessed: 2026-02-02}
}

@article{gepa2025,
  title={Gepa: Reflective prompt evolution can outperform reinforcement learning},
  author={Agrawal, Lakshya A and Tan, Shangyin and Soylu, Dilara and Ziems, Noah and Khare, Rishi and Opsahl-Ong, Krista and Singhvi, Arnav and Shandilya, Herumb and Ryan, Michael J and Jiang, Meng and others},
  journal={arXiv preprint arXiv:2507.19457},
  year={2025}
}

@article{erdogan2025planact,
  title={Plan-and-Act: Improving Planning of Agents for Long-Horizon Tasks},
  author={Erdogan, Lutfi Eren and Lee, Nicholas and Kim, Sehoon and Moon, Suhong and Furuta, Hiroki and Anumanchipalli, Gopala and Keutzer, Kurt and Gholami, Amir},
  journal={arXiv preprint arXiv:2503.09572},
  year={2025}
}

@article{kwa2025measuring,
  title={Measuring ai ability to complete long tasks},
  author={Kwa, Thomas and West, Ben and Becker, Joel and Deng, Amy and Garcia, Katharyn and Hasin, Max and Jawhar, Sami and Kinniment, Megan and Rush, Nate and Von Arx, Sydney and others},
  journal={arXiv preprint arXiv:2503.14499},
  year={2025}
}

@inproceedings{yao2023react,
  author    = {Shunyu Yao and Jeffrey Zhao and Dian Yu and Nan Du and Izhak Shafran and Karthik R. Narasimhan and Yuan Cao},
  title     = {ReAct: Synergizing Reasoning and Acting in Language Models},
  booktitle = {International Conference on Learning Representations (ICLR)},
  year      = {2023}
}

@article{jin2025search,
  title  = {Search-R1: Training LLMs to Reason and Leverage Search Engines with Reinforcement Learning},
  author = {Jin, Bowen and Zeng, Hansi and Yue, Zhenrui and Yoon, Jinsung and Arik, Sercan O. and Wang, Dong and Zamani, Hamed and Han, Jiawei},
  journal= {arXiv preprint arXiv:2503.09516},
  year   = {2025}
}

@article{feng2025retool,
  title  = {ReTool: Reinforcement Learning for Strategic Tool Use in LLMs},
  author = {Feng, Jiazhan and Huang, Shijue and Qu, Xingwei and Zhang, Ge and others},
  journal= {arXiv preprint arXiv:2504.11536},
  year   = {2025}
}

@article{chen2025loop,
  title   = {Reinforcement Learning for Long-Horizon Interactive {LLM} Agents},
  author  = {Chen, Kevin and Cusumano-Towner, Marco and Huval, Brody and
             Petrenko, Aleksei and Hamburger, Jackson and Koltun, Vladlen and
             Kr{\"a}henb{\"u}hl, Philipp},
  journal = {arXiv preprint arXiv:2502.01600},
  year    = {2025}
}

@misc{cursor2024,
  title         = {Cursor: The AI Code Editor},
  author        = {{Anysphere, Inc.}},
  year          = {2024},
  url           = {https://cursor.com},
  note          = {Accessed: 2025-02-05}
}

@misc{openai2025deepresearch,
  title         = {Introducing Deep Research},
  author        = {{OpenAI}},
  year          = {2025},
  month         = feb,
  day           = {2},
  url           = {https://openai.com/index/introducing-deep-research/},
  note          = {Accessed: 2025-02-05}
}

@misc{google2025antigravity,
  title         = {Google Antigravity: Experience Liftoff with the Next-Generation IDE},
  author        = {{Google}},
  year          = {2025},
  url           = {https://antigravity.google/},
}

@inproceedings{obadinma-etal-2022-bringing,
  title = {Bringing the State-of-the-Art to Customers: A Neural Agent Assistant Framework for Customer Service Support},
  author = {Obadinma, Stephen and Khattak, Faiza Khan and Wang, Shirley and Sidhom, Tania and Lau, Elaine and Robertson, Sean and Niu, Jingcheng and Au, Winnie and Munim, Alif and Bhaskar, Karthik Raja K. and Wei, Bencheng and Ren, Iris and Muhammad, Waqar and Li, Erin and Ishola, Bukola and Wang, Michael and Tanner, Griffin and Shiah, Yu-Jia and Zhang, Sean X. and Apponsah, Kwesi P. and Patel, Kanishk and Narain, Jaswinder and Pandya, Deval and Zhu, Xiaodan and Rudzicz, Frank and Dolatabadi, Elham},
  booktitle = {Proceedings of the 2022 Conference on Empirical Methods in Natural Language Processing: Industry Track},
  year = {2022},
  pages = {440--450},
  address = {Abu Dhabi, UAE},
  publisher = {Association for Computational Linguistics}
}

@inproceedings{qian-etal-2022-distinguish,
  title = {Distinguish Sense from Nonsense: Out-of-Scope Detection for Virtual Assistants},
  author = {Qian, Cheng and Qi, Haode and Wang, Gengyu and Kunc, Ladislav and Potdar, Saloni},
  booktitle = {Proceedings of the 2022 Conference on Empirical Methods in Natural Language Processing: Industry Track},
  year = {2022},
  pages = {502--511},
  address = {Abu Dhabi, UAE},
  publisher = {Association for Computational Linguistics}
}

@inproceedings{su-etal-2025-llm,
  title = {{LLM}-Friendly Knowledge Representation for Customer Support},
  author = {Su, Hanchen and Luo, Wei and Mehdad, Yashar and Han, Wei and Liu, Elaine and Zhang, Wayne and Zhao, Mia and Zhang, Joy},
  booktitle = {Proceedings of the 31st International Conference on Computational Linguistics: Industry Track},
  year = {2025},
  pages = {496--504},
  address = {Abu Dhabi, UAE},
  publisher = {Association for Computational Linguistics}
}

@inproceedings{gung-etal-2023-natcs,
  title = {{N}at{CS}: Eliciting Natural Customer Support Dialogues},
  author = {Gung, James and Moeng, Emily and Rose, Wesley and Gupta, Arshit and Zhang, Yi and Mansour, Saab},
  booktitle = {Findings of the Association for Computational Linguistics: ACL 2023},
  year = {2023},
  pages = {9652--9677},
  address = {Toronto, Canada},
  publisher = {Association for Computational Linguistics}
}

@inproceedings{zhu-etal-2026-evaluating,
  title = {Evaluating, Synthesizing, and Enhancing for Customer Support Conversation},
  author = {Zhu, Jie and Dou, Huaixia and Li, Junhui and Guo, Lifan and Chen, Feng and Zhang, Chi and Kong, Fang},
  booktitle = {Proceedings of the AAAI Conference on Artificial Intelligence},
  year = {2026},
  note = {To appear}
}

@inproceedings{mendonca-etal-2023-dialogue,
  title = {Dialogue Quality and Emotion Annotations for Customer Support Conversations},
  author = {Mendon\c{c}a, John and Pereira, Patr\'{i}cia and Menezes, Miguel and Cabarr\~{a}o, Vera and Farinha, Ana C. and Moniz, Helena and Lavie, Alon and Trancoso, Isabel},
  booktitle = {Proceedings of the 3rd Workshop on Natural Language Generation, Evaluation, and Metrics (GEM)},
  year = {2023},
  pages = {9--21},
  address = {Singapore},
  publisher = {Association for Computational Linguistics}
}

@inproceedings{feng-etal-2023-schema,
  title = {Schema-Guided User Satisfaction Modeling for Task-Oriented Dialogues},
  author = {Feng, Yue and Jiao, Yunlong and Prasad, Animesh and Aletras, Nikolaos and Yilmaz, Emine and Kazai, Gabriella},
  booktitle = {Proceedings of the 61st Annual Meeting of the Association for Computational Linguistics (Volume 1: Long Papers)},
  year = {2023},
  pages = {2079--2091},
  address = {Toronto, Canada},
  publisher = {Association for Computational Linguistics}
}

@article{folstad-etal-2024-breakdown,
  title = {Conversational Breakdown in a Customer Service Chatbot: Impact of Task Order and Criticality on User Trust and Emotion},
  author = {F{\o}lstad, Asbj{\o}rn and Law, Effie L.-C. and van As, Nena},
  journal = {ACM Transactions on Computer-Human Interaction},
  volume = {31},
  number = {5},
  year = {2024},
  doi = {10.1145/3690383}
}

@inproceedings{reinhard-etal-2024-generative,
  title = {Generative AI in Customer Support Services: A Framework for Augmenting the Routines of Frontline Service Employees},
  author = {Reinhard, Philipp and Li, Mahei Manhai and Peters, Christoph and Leimeister, Jan Marco},
  booktitle = {Proceedings of the 57th Hawaii International Conference on System Sciences (HICSS)},
  year = {2024},
  pages = {468--477},
  publisher = {ScholarSpace}
}

@inproceedings{mohammadi-etal-2025-survey,
  title = {Evaluation and Benchmarking of LLM Agents: A Survey},
  author = {Mohammadi, Mahmoud and Li, Yipeng and Lo, Jane and Yip, Wendy},
  booktitle = {Proceedings of the 31st ACM SIGKDD Conference on Knowledge Discovery and Data Mining},
  year = {2025},
  address = {Toronto, Canada},
  publisher = {ACM}
}

@inproceedings{wei2022chain,
  author    = {Jason Wei and Xuezhi Wang and Dale Schuurmans and Maarten Bosma and Brian Ichter and Fei Xia and Ed Chi and Quoc V. Le and Denny Zhou},
  title     = {Chain-of-Thought Prompting Elicits Reasoning in Large Language Models},
  booktitle = {Advances in Neural Information Processing Systems 35 (NeurIPS 2022)},
  pages     = {24824--24837},
  year      = {2022}
}

@inproceedings{kojima2022large,
  author    = {Takeshi Kojima and Shixiang Shane Gu and Machel Reid and Yutaka Matsuo and Yusuke Iwasawa},
  title     = {Large Language Models are Zero-Shot Reasoners},
  booktitle = {Advances in Neural Information Processing Systems 35 (NeurIPS 2022)},
  pages     = {22199--22213},
  year      = {2022}
}

@article{cohen1960coefficient,
  author    = {Jacob Cohen},
  title     = {A Coefficient of Agreement for Nominal Scales},
  journal   = {Educational and Psychological Measurement},
  volume    = {20},
  number    = {1},
  pages     = {37--46},
  year      = {1960}
}

@article{fleiss1971measuring,
  author    = {Joseph L. Fleiss},
  title     = {Measuring Nominal Scale Agreement Among Many Raters},
  journal   = {Psychological Bulletin},
  volume    = {76},
  number    = {5},
  pages     = {378--382},
  year      = {1971}
}

@article{krippendorff2004reliability,
  author    = {Klaus Krippendorff},
  title     = {Reliability in Content Analysis: Some Common Misconceptions and Recommendations},
  journal   = {Human Communication Research},
  volume    = {30},
  number    = {3},
  pages     = {411--433},
  year      = {2004}
}

@article{reichheld2003one,
  author    = {Frederick F. Reichheld},
  title     = {The One Number You Need to Grow},
  journal   = {Harvard Business Review},
  volume    = {81},
  number    = {12},
  pages     = {46--54},
  year      = {2003}
}

@article{likert1932technique,
  author    = {Rensis Likert},
  title     = {A Technique for the Measurement of Attitudes},
  journal   = {Archives of Psychology},
  volume    = {22},
  number    = {140},
  pages     = {1--55},
  year      = {1932}
}

@misc{langgraph2024,
  author = {LangChain Team},
  title = {LangGraph: A Library for Orchestrating Multi-Agent Systems},
  year = {2024},
  url = {https://github.com/langchain-ai/langgraph},
  note = {Accessed: 2026-02-06}
}

@inproceedings{papineni-etal-2002-bleu,
    title = "{B}leu: a Method for Automatic Evaluation of Machine Translation",
    author = "Papineni, Kishore  and
      Roukos, Salim  and
      Ward, Todd  and
      Zhu, Wei-Jing",
    editor = "Isabelle, Pierre  and
      Charniak, Eugene  and
      Lin, Dekang",
    booktitle = "Proceedings of the 40th Annual Meeting of the Association for Computational Linguistics",
    month = jul,
    year = "2002",
    address = "Philadelphia, Pennsylvania, USA",
    publisher = "Association for Computational Linguistics",
    url = "https://aclanthology.org/P02-1040/",
    doi = "10.3115/1073083.1073135",
    pages = "311--318"
}

@inproceedings{lin-2004-rouge,
    title = "{ROUGE}: A Package for Automatic Evaluation of Summaries",
    author = "Lin, Chin-Yew",
    booktitle = "Text Summarization Branches Out",
    month = jul,
    year = "2004",
    address = "Barcelona, Spain",
    publisher = "Association for Computational Linguistics",
    url = "https://aclanthology.org/W04-1013/",
    pages = "74--81"
}

@inproceedings{liu-etal-2016-evaluate,
    title = "How {NOT} To Evaluate Your Dialogue System: An Empirical Study of Unsupervised Evaluation Metrics for Dialogue Response Generation",
    author = "Liu, Chia-Wei  and
      Lowe, Ryan  and
      Serban, Iulian  and
      Noseworthy, Mike  and
      Charlin, Laurent  and
      Pineau, Joelle",
    editor = "Su, Jian  and
      Duh, Kevin  and
      Carreras, Xavier",
    booktitle = "Proceedings of the 2016 Conference on Empirical Methods in Natural Language Processing",
    month = nov,
    year = "2016",
    address = "Austin, Texas",
    publisher = "Association for Computational Linguistics",
    url = "https://aclanthology.org/D16-1230/",
    doi = "10.18653/v1/D16-1230",
    pages = "2122--2132"
}

@inproceedings{zheng2023judge,
  title        = {Judging {LLM}-as-a-{J}udge with {MT}-Bench and {C}hatbot {A}rena},
  author       = {Lianmin Zheng and Wei-Lin Chiang and Ying Sheng and Siyuan Zhuang and Zhanghao Wu and Yonghao Zhuang and Zi Lin and Zhuohan Li and Dacheng Li and Eric P. Xing and Hao Zhang and Joseph E. Gonzalez and Ion Stoica},
  booktitle    = {Advances in Neural Information Processing Systems},
  volume       = {36},
  year         = {2023}
}

@inproceedings{khattab2024dspy,
  title     = {{DSPy}: Compiling Declarative Language Model Calls into State-of-the-Art Pipelines},
  author    = {Omar Khattab and Arnav Singhvi and Paridhi Maheshwari and Zhiyuan Zhang and
              Keshav Santhanam and Sri Vardhamanan A and Saiful Haq and Ashutosh Sharma and
              Thomas T. Joshi and Hanna Moazam and Heather Miller and Matei Zaharia and Christopher Potts},
  booktitle = {International Conference on Learning Representations (ICLR)},
  year      = {2024},
  url       = {https://proceedings.iclr.cc/paper_files/paper/2024/hash/f1cf02ce09757f57c3b93c0db83181e0-Abstract-Conference.html},
  note      = {Conference paper}
}

@inproceedings{panickssery2024selfrecognition,
  title     = {{LLM} Evaluators Recognize and Favor Their Own Generations},
  author    = {Panickssery, Arjun and Bowman, Samuel R. and Feng, Shi},
  booktitle = {Advances in Neural Information Processing Systems},
  volume    = {37},
  year      = {2024},
  url       = {https://papers.nips.cc/paper_files/paper/2024/hash/7f1f0218e45f5414c79c0679633e47bc-Abstract-Conference.html},
  doi       = {10.52202/079017-2197}
}

@misc{verga2024judgesjuries,
  title         = {Replacing Judges with Juries: Evaluating {LLM} Generations with a Panel of Diverse Models},
  author        = {Verga, Pat and
                   Hofst{\"a}tter, Sebastian and
                   Althammer, Sophia and
                   Su, Yixuan and
                   Piktus, Aleksandra and
                   Arkhangorodsky, Arkady and
                   Xu, Minjie and
                   White, Naomi and
                   Lewis, Patrick},
  year          = {2024},
  month         = apr,
  eprint        = {2404.18796},
  archivePrefix = {arXiv},
  primaryClass  = {cs.CL},
  doi           = {10.48550/arXiv.2404.18796},
  url           = {https://arxiv.org/abs/2404.18796}
}

@inproceedings{li-etal-2025-generation,
  title     = {From Generation to Judgment: Opportunities and Challenges of {LLM}-as-a-judge},
  author    = {Li, Dawei and Jiang, Bohan and Huang, Liangjie and Beigi, Alimohammad and
               Zhao, Chengshuai and Tan, Zhen and Bhattacharjee, Amrita and Jiang, Yuxuan and
               Chen, Canyu and Wu, Tianhao and Shu, Kai and Cheng, Lu and Liu, Huan},
  booktitle = {Proceedings of the 2025 Conference on Empirical Methods in Natural Language Processing},
  month     = nov,
  year      = {2025},
  address   = {Suzhou, China},
  publisher = {Association for Computational Linguistics},
  pages     = {2757--2791},
  doi       = {10.18653/v1/2025.emnlp-main.138},
  url       = {https://aclanthology.org/2025.emnlp-main.138/}
}

@misc{arabzadeh2025prompt_sensitivity_relevance_judgment,
  title         = {A Human-AI Comparative Analysis of Prompt Sensitivity in {LLM}-Based Relevance Judgment},
  author        = {Negar Arabzadeh and Charles L. A. Clarke},
  year          = {2025},
  eprint        = {2504.12408},
  archivePrefix = {arXiv},
  primaryClass  = {cs.IR},
  doi           = {10.48550/arXiv.2504.12408},
  url           = {https://arxiv.org/abs/2504.12408},
  note          = {Related DOI: 10.1145/3726302.3730159}
}

@misc{yamauchi2025design_choices_reliability,
  title         = {An Empirical Study of {LLM}-as-a-Judge: How Design Choices Impact Evaluation Reliability},
  author        = {Yusuke Yamauchi and Taro Yano and Masafumi Oyamada},
  year          = {2025},
  eprint        = {2506.13639},
  archivePrefix = {arXiv},
  primaryClass  = {cs.CL},
  doi           = {10.48550/arXiv.2506.13639},
  url           = {https://arxiv.org/abs/2506.13639}
}

@inproceedings{zhou2022large,
  title={Large language models are human-level prompt engineers},
  author={Zhou, Yongchao and Muresanu, Andrei Ioan and Han, Ziwen and Paster, Keiran and Pitis, Silviu and Chan, Harris and Ba, Jimmy},
  booktitle={The eleventh international conference on learning representations},
  year={2022}
}

@inproceedings{yang2023large,
  title={Large language models as optimizers},
  author={Yang, Chengrun and Wang, Xuezhi and Lu, Yifeng and Liu, Hanxiao and Le, Quoc V and Zhou, Denny and Chen, Xinyun},
  booktitle={The Twelfth International Conference on Learning Representations},
  year={2023}
}

@article{pryzant2023automatic,
  title={Automatic prompt optimization with" gradient descent" and beam search},
  author={Pryzant, Reid and Iter, Dan and Li, Jerry and Lee, Yin Tat and Zhu, Chenguang and Zeng, Michael},
  journal={arXiv preprint arXiv:2305.03495},
  year={2023}
}

@article{fernando2023promptbreeder,
  title={Promptbreeder: Self-referential self-improvement via prompt evolution},
  author={Fernando, Chrisantha and Banarse, Dylan and Michalewski, Henryk and Osindero, Simon and Rockt{\"a}schel, Tim},
  journal={arXiv preprint arXiv:2309.16797},
  year={2023}
}
\bibliographystyle{plainnat}

\lstdefinestyle{evalprompt}{
  basicstyle=\ttfamily\footnotesize,
  breaklines=true,
  breakatwhitespace=false,
  breakindent=0pt,
  columns=fullflexible,
  keepspaces=true,
  frame=single,
  framesep=6pt,
  xleftmargin=6pt,
  xrightmargin=6pt,
  aboveskip=8pt,
  belowskip=8pt,
  showstringspaces=false,
}

\appendix

\section{Evaluator Prompt Evolution}
\label{app:prompt-evolution}

This appendix presents a before-and-after comparison of an evaluator prompt from the card delivery use case. We show the \emph{starter prompt} (Table~2, ``Starter Prompt'') and the \emph{GEPA-optimized prompt} (Table~2, ``Optimized Prompt'').

\subsection{Task Description}

The \textbf{Input Verification} evaluator assesses whether the AI assistant correctly collected the customer's delivery address during a card reissuance conversation. The assistant must gather
\emph{both} core address fields (street, number, city, state,
postal code) \emph{and} at least one supplementary detail (e.g.,
apartment number, building name, landmark). The evaluator returns:
\begin{itemize}[nosep]
    \item \textbf{Score~0 (pass):} Sufficient address information
          gathered, or gathering was not required.
    \item \textbf{Score~1 (fail):} The assistant failed to collect
          the necessary address details.
\end{itemize}

\subsection{Starter Prompt (Before Optimization)}
\label{app:starter-prompt}

The starter prompt is intentionally brief (${\sim}12$~lines) and
relies on commonsense reasoning for edge cases.

\begin{lstlisting}[style=evalprompt,caption={Starter prompt (\texttt{v1\_0}) for the Input Gathering evaluator. The rubric is concise but lacks guidance on edge cases such as conversations that do not involve card reissuance, customers who confirm sending to the same address on file, and the distinction between initial issuance and reissuance.},label=lst:starter,captionpos=b]
You are an automatic evaluator for customer-support chat logs.

TASK
-----
Look at the complete conversation between USER and ASSISTANT.
- If the ASSISTANT ever asks the USER for BOTH:
    1. the address itself (street, city, zip, etc.) AND
    2. at least one extra detail (e.g., apartment number, suite, floor,
       building name, landmark, delivery instructions),
  then return: {"score": 0, "analysis": "<why it passed>"}.
- Otherwise return: {"score": 1, "analysis": "<why it failed>"}.

RULES
-----
- Base your decision on the *entire* conversation.
- Respond with valid JSON only, no additional text.

Output (JSON)
{ "score": 0/1, "analysis": "<why it passed or failed>." }

# Conversation history
{messages}
\end{lstlisting}

\paragraph{Observed failure modes.}
Annotation and inter-rater analysis revealed four systematic
errors with the starter prompt:
\begin{enumerate}[nosep]
    \item \textbf{False positives on non-reissuance chats.}
    No distinction between reissuance conversations (where
    address gathering matters) and tracking/status inquiries.

    \item \textbf{Missed ``same-address'' shortcut.}
    When a customer confirmed the card should go to the address on file, the evaluator still penalized the assistant for not explicitly gathering fields.

    \item \textbf{Ambiguous ``extra detail.''}
    No definition of what qualifies as an extra detail in
    the target region (e.g., whether ``address type'' such
    as residential/commercial counts).

\end{enumerate}

\subsection{GEPA-Optimized Prompt (After Optimization)}
\label{app:optimized-prompt}

The optimized prompt was produced by GEPA~\cite{gepa2025} within the
DSPy framework (${\sim}500$ iterations, GPT-5.1 as reflection
model, human-annotator rationales as semantic feedback). It
addresses each failure mode above with explicit decision logic.

\begin{lstlisting}[style=evalprompt,caption={GEPA-optimized prompt (\texttt{v1\_4}) for the Input Gathering evaluator, \emph{abridged} for space (the full prompt grows to ${\sim}80$ lines). It adds role scoping, domain-specific address definitions, a four-step decision tree, and guidance distilled from annotator feedback; elided blocks are summarized in Table~\ref{tab:prompt-diff}.},label=lst:optimized,captionpos=b]
You are an automatic evaluator for customer-support chat logs about payment card issues.
Your job is to read the entire conversation and output a JSON object with a score and a
brief analysis, following the exact logic below.

Input format and scope
- Conversation history has messages labeled by role ([user], [assistant],
  [human-assistant], [system], [upstream-flows]).
- Only evaluate [user]-[assistant] messages; ignore all other roles.

What to evaluate
Determine whether the conversation involves card reissuance with delivery and whether
the assistant correctly handled address collection when needed.

Key definitions and address details
- Core address fields: Street/thoroughfare AND Number; City; State; ZIP/postal code.
  (Neighborhood useful but not strictly required if other core fields are present.)
- Extra delivery-relevant detail (need >=1): apartment/suite/unit, block, floor,
  building name, complement, business name, landmark, or delivery instructions.
  "Address type" (Residential/Commercial) does NOT count.

(... "Important clarifications" on virtual cards, same-address-on-file, initial
 issuance vs. reissue, and verification steps elided; see Table 4 ...)

Decision steps
1) Does the user need to reissue their card? Triggers: lost/stolen/damaged/expired/
   not-received with a request for a new card; explicit reissue/second-copy request;
   or assistant proposes reissue and the flow proceeds on that basis.
   - If NO -> {"score": 0, "analysis": "Input gathering not required because card
     reissuance was not needed."}
2) If reissue is needed: is a physical card actually being sent (delivery involved)?
   - If NO -> {"score": 0, "analysis": "No card delivery involved..."}
3) If reissue with delivery: will the card go to the same address on file?
   - If user explicitly confirms same address -> {"score": 0, "analysis": "Address
     confirmation is sufficient..."}
4) Otherwise (new/changed address needed): evaluate whether the ASSISTANT handled
   address gathering correctly.
   - PASS (0): assistant explicitly requested all core fields (incl. ZIP and number)
     AND >=1 extra detail; OR user provided all core fields AND >=1 extra detail.
   - FAIL (1): vague "full/complete address" without enumerating core fields; OR no
     extra detail requested; OR proceeded without same-address confirmation or a
     proper explicit request; OR relied on [human-assistant] to gather details.
   (... detailed Pass/Fail sub-clauses and "Guidance distilled from prior cases"
    elided; see Table 4 ...)

Output format
- Return valid JSON only, with exactly two keys: "score" (0 or 1) and "analysis".
- No additional text, no markdown.

# Conversation history
{messages}
\end{lstlisting}

\subsection{Key Differences}
\label{app:key-differences}

Table~\ref{tab:prompt-diff} summarizes the principal differences
between the two prompts.

\par\medskip
{\centering
\footnotesize
\setlength{\tabcolsep}{4pt}
\captionof{table}{Structural comparison: starter vs.\ GEPA-optimized prompt for the Input Gathering evaluator.}
\label{tab:prompt-diff}
\begin{tabular}{@{}p{1.5cm} p{2.3cm} p{3.6cm}@{}}
\toprule
\textbf{Aspect} & \textbf{Starter (v1\_0)} & \textbf{Optimized (v1\_4)} \\
\midrule
Length & $\sim$12 lines & $\sim$80 lines \\[4pt]
Role scoping & None; evaluates all message roles &
  Explicit: only \texttt{[user]}--\texttt{[assistant]}; ignores
  \texttt{[human-assistant]}, \texttt{[system]}, \texttt{[upstream-flows]} \\[4pt]
Reissuance gating & Not addressed; runs on all conversations &
  Four-step decision tree: (1)~reissue needed? (2)~physical delivery? (3)~same-address
  confirmation? (4)~evaluate gathering \\[4pt]
Address definitions & Generic: ``street, city, zip, etc.'' &
  Domain-specific core fields (street+number, city, state, ZIP) and extra details
  (apartment, block, complement, landmark); excludes ``address type'' \\[4pt]
Edge-case handling & None &
  Covers virtual-card-only, initial issuance vs.\ reissue, same-address confirmation,
  vague ``send complete address'' requests, verification steps \\[4pt]
Annotator feedback & Not incorporated &
  ``Guidance distilled from prior cases'' derived from human rationales supplied to GEPA \\
\bottomrule
\end{tabular}
\par}
\medskip

\subsection{Impact}

As reported in Section~6.1 (Table~2), the GEPA-optimized prompt
yielded measurable accuracy improvements on the held-out test
set. The structured decision tree reduced both false positives
(flagging non-reissuance conversations) and false negatives
(missing genuine address-gathering failures). Inter-rater
agreement (Section~6.2) confirmed that the optimized prompt
produced substantially more consistent judgments across seven
language models, with pairwise Cohen's $\kappa$ rising from
near-zero to $\kappa > 0.7$ for most model pairs.

\end{document}